\documentclass[sigconf]{acmart}

\AtBeginDocument{
  \providecommand\BibTeX{{%
    \normalfont B\kern-0.5em{\scshape i\kern-0.25em b}\kern-0.8em\TeX}}}

\setcopyright{cc}
\setcctype{by}
\copyrightyear{2026}
\acmYear{2026}
\acmDOI{10.1145/3770855.3817581}
\acmConference[KDD 2026]{Proceedings of the 32nd ACM SIGKDD Conference on Knowledge Discovery and Data Mining V.2}{August 9--13, 2026}{Jeju Island, Republic of Korea.}
\acmBooktitle{Proceedings of the 32nd ACM SIGKDD Conference on Knowledge Discovery and Data Mining V.2 (KDD 2026), August 9--13, 2026, Jeju Island, Republic of Korea}
\acmISBN{979-8-4007-2259-2/2026/08}

\usepackage{algorithm}
\usepackage{algpseudocode}
\usepackage{amsmath}
\usepackage{multirow}
\usepackage{graphicx}
\usepackage{xspace}
\usepackage{colortbl}
\usepackage{xcolor}
\usepackage{booktabs}
\usepackage{enumitem}
\usepackage{balance}
\usepackage{pifont}
\usepackage[most]{tcolorbox}
\definecolor{proprietary}{RGB}{255,235,235}
\definecolor{opensource}{RGB}{235,255,235}
\definecolor{baselinegray}{RGB}{240,240,248}
\definecolor{oursgreen}{RGB}{232,247,232}
\definecolor{ablationtan}{RGB}{248,243,232}

\newcommand{\bench}{\textsc{SafeBuild\allowbreak-\allowbreak Bench}\xspace}
\newcommand{\method}{\textbf{GEMS}\xspace}
\newcommand{\safebuildcheck}{\ding{51}}

\begin{document}

\title{\bench: A Temporal-Robust Construction Safety Benchmark with Graph-Enhanced Data Mining}

\author{Yi Cui}
\orcid{0009-0002-2447-4422}
\authornote{Yi Cui and Zilin Wang contributed equally to this research.}
\affiliation{
  \institution{Hong Kong University of Science and Technology (Guangzhou)}
  \city{Guangzhou}
  \country{China}
}
\email{ycui785@connect.hkust-gz.edu.cn}

\author{Zilin Wang}
\orcid{0009-0005-4865-1050}
\authornotemark[1]
\affiliation{
  \institution{The Hong Kong University of Science and Technology(Guangzhou)}
  \city{Guangzhou}
  \country{China}
}
\email{zwang374@connect.hkust-gz.edu.cn}

\author{Yijie Xu}
\orcid{0009-0008-4529-2701}
\affiliation{
  \institution{Hong Kong University of Science and Technology (Guangzhou)}
  \city{Guangzhou}
  \country{China}
}
\email{yxu409@connect.hkust-gz.edu.cn}

\author{Qianyi Cai}
\orcid{0009-0007-0990-5649}
\affiliation{
  \institution{The Hong Kong University of Science and Technology(Guangzhou)}
  \city{Guangzhou}
  \country{China}
}
\email{qcai603@connect.hkust-gz.edu.cn}

\author{Huizai Yao}
\orcid{0009-0007-5245-349X}
\affiliation{
  \institution{The Hong Kong University of Science and Technology(Guangzhou)}
  \city{Guangzhou}
  \country{China}
}
\email{huizai.yao@connect.hkust-gz.edu.cn}

\author{Shuai Jiang}
\orcid{0000-0003-4413-6914}
\affiliation{
  \department{School of Management}
  \institution{Northwest Polytechnical University Xi'an}
  \city{Xi'an}
  \country{China}
}
\email{shuaijiangai@gmail.com}

\author{Bingzhuo Zhong}
\orcid{0000-0001-6557-2374}
\authornote{Bingzhuo Zhong and Hui Xiong are corresponding authors.}
\affiliation{
  \department{The Thrust of Artificial Intelligence, Information Hub}
  \institution{The Hong Kong University of Science and Technology(Guangzhou)}
  \city{Guangzhou}
  \country{China}
}
\affiliation{
  \department{The Thrust of Intelligent Transportation, System Hub}
  \institution{The Hong Kong University of Science and Technology(Guangzhou)}
  \city{Guangzhou}
  \country{China}
}
\email{bingzhuoz@hkust-gz.edu.cn}

\author{Hui Xiong}
\orcid{0000-0001-6016-6465}
\authornotemark[2]
\affiliation{
  \department{Thrust of Artificial Intelligence}
  \institution{The Hong Kong University of Science and Technology(Guangzhou)}
  \city{Guangzhou}
  \country{China}
}
\affiliation{
  \department{Department of Computer Science and Engineering}
  \institution{The Hong Kong University of Science and Technology}
  \city{Hong Kong SAR}
  \country{Hong Kong}
}
\email{xionghui@ust.hk}

\renewcommand{\shortauthors}{Yi Cui et al.}

\begin{abstract}
Construction-safety models must handle concrete deployment risks, such as a worker standing near a scaffold edge without guardrails, rather than only recognize common objects in curated images. Yet real inspection archives are redundant, long-tailed, and collected across changing sites and months. We introduce \bench, a metadata-driven benchmark for evaluating multimodal large language models on construction safety under realistic temporal and site variation. It is mined from 100K+ industrial image--text records and contains 3,314 task instances from over 3,000 expert-verified images, covering multiple-choice hazard identification and free-form hazard description. To make expert verification scalable, we develop \method, a graph-enhanced multimodal selection pipeline that combines a proxy-model confusion signal with graph-based diversity to identify informative candidates from redundant streams. On public instruction-tuning data, \method-selected subsets preserve robustness-oriented performance under small data budgets. On \bench, current MLLMs remain far from reliable construction-safety understanding, with the best overall score near 60. We release the benchmark, evaluation scripts, and \method codebase at \url{https://github.com/safebuild/gems}.
\end{abstract}

\ccsdesc[500]{Computing methodologies~Artificial intelligence}
\ccsdesc[300]{Information systems~Data mining}

\keywords{Benchmark, Data-Centric AI, Temporal Robustness, Data Efficiency, Construction Safety}

\maketitle

\section{Introduction}\label{sec:intro}

Multimodal Large Language Models (MLLMs) have improved general-purpose visual understanding and instruction following~\cite{gpt4o_system_card,qwen25_vl,llava15,gemini25}, but their reliability in safety-critical domains is still uncertain. Industrial safety monitoring spans multiple settings, where failures can cause injuries, downtime, and compliance risks. In construction, for example, a model must distinguish an orderly scaffold from a scaffold-edge scene without guardrails, under clutter, changing weather, and evolving site layouts. We study this setting because such reliability questions directly affect deployment decisions.

\begin{figure}[b]
    \centering
    \includegraphics[width=\linewidth]{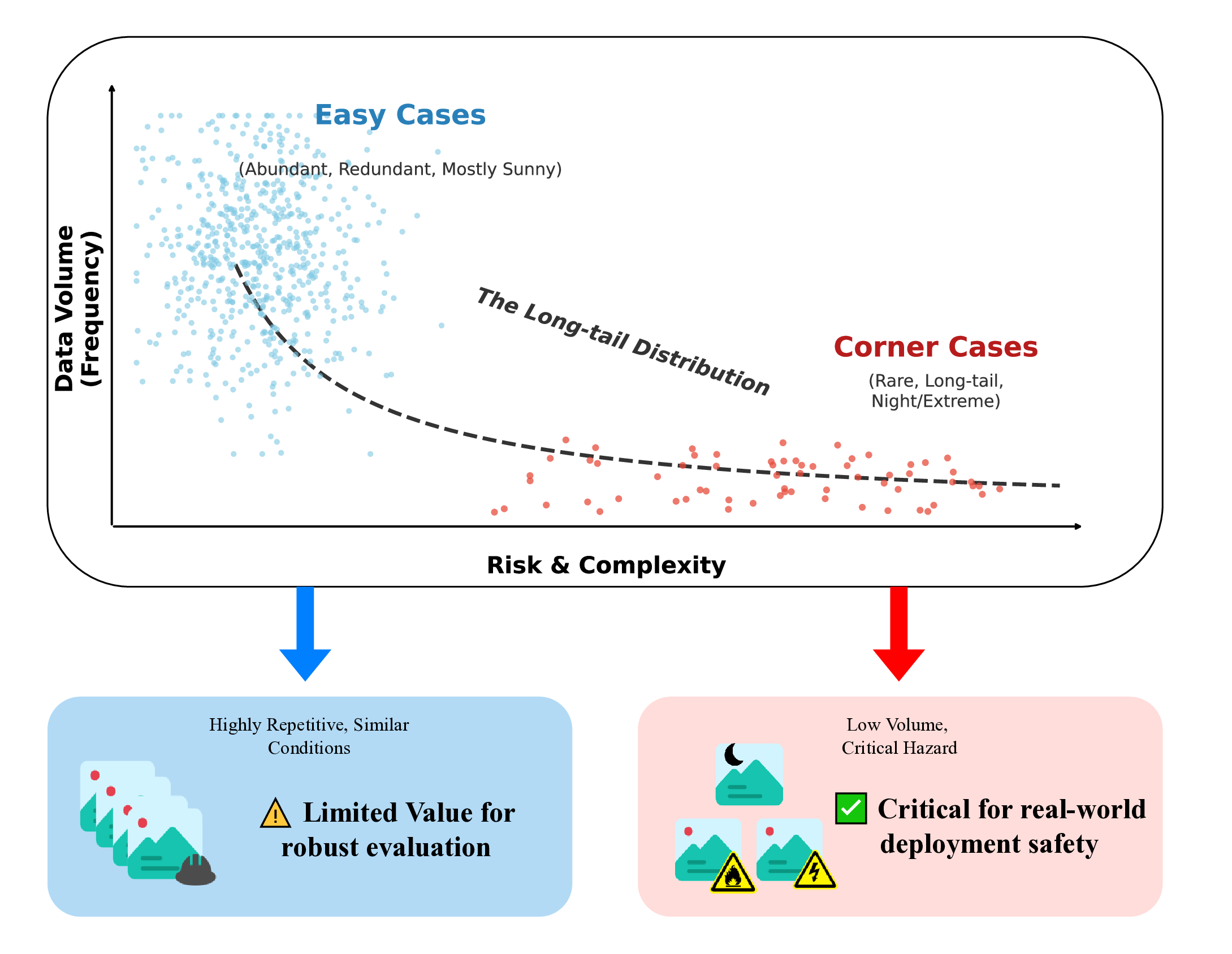}
    \caption{\textbf{The dichotomy of construction safety data.}
    \textbf{(Top)} Abundant data (blue) is repetitive and low-risk, whereas critical threats form a sparse, complex long tail (red).
    \textbf{(Bottom)} ``Easy cases'' dominate volume, while rare corner cases are often missed yet drive real-world safety incidents.}
    \Description{A two-part illustration showing that common construction-site records are repetitive and low-risk, while rare hazardous cases form a sparse long tail.}
    \label{fig:intro_dichotomy}
    \vspace{-1em}
\end{figure}

Construction safety monitoring produces archives of inspection records in the form of image--text pairs. Despite their size, these archives are often data-rich but information-poor: much of the volume is repetitive and low-risk, while rare but high-impact hazards form a long tail~\cite{DBLP:journals/ijcv/YangJSG22}. Figure~\ref{fig:intro_dichotomy} illustrates this imbalance. Meanwhile, construction sites evolve over time due to weather, construction progress, camera maintenance, and lighting, producing temporal distribution shift~\cite{DBLP:conf/nips/YaoCC0KF22}. Evaluation protocols that ignore this structure can overestimate deployment performance, especially when models exploit stable backgrounds or site-specific cues rather than safety-relevant semantics. A useful benchmark should therefore retain time and site metadata so that users can stratify performance by deployment-relevant slices instead of relying only on an aggregate score.

These gaps raise a practical question: \textit{Can we build a construction-safety benchmark that (1) enables temporal and site-stratified evaluation, (2) concentrates on long-tail hazards instead of redundant data, and (3) remains feasible to build from large-scale data streams?} 
To address this, we introduce \textbf{\bench}, a construction safety benchmark mined from 100{,}000+ raw image--text pairs collected across multiple sites and dates. The release contains 3,314 task instances from over 3,000 expert-verified images. Each instance retains temporal and site metadata, enabling users to analyze performance by time period and site identity rather than relying on a single fixed split. \bench covers hierarchical tasks ranging from hazard identification to hazard description, with executable evaluation scripts and a standardized LLM-as-a-judge scheme for scalable free-form scoring.

Building such a benchmark from redundant streams requires scalable candidate mining before expert verification. We therefore develop an automated curation pipeline, \method (\underline{G}raph-\underline{E}nhanced \underline{M}ultimodal \underline{S}election), to extract a small set of informative and diverse candidates from large archives. \method combines epistemic uncertainty, which ranks samples where current MLLMs are unsure, with a dual graph structure that enforces coverage across scenes and reduces near duplicates. The result is an \emph{informational core}: a compact subset that (i) preserves coverage of the major scene modes and hazard categories, (ii) concentrates hard and rare hazards that are otherwise diluted by repetition, and (iii) removes repeated or near-identical frames that contribute little new information.

We validate the general utility of \method with a proxy study on a public instruction-tuning corpus. Fine-tuning LLaVA-v1.5 on only the top 1\% \method-selected subset, about 6.6K samples on LLaVA-mix-665K, matches or exceeds full-data training on robustness-oriented benchmarks. This result supports the use of \method as a practical mining tool for benchmark construction, since it removes redundancy while retaining high-value long-tail content.

Across a diverse set of MLLMs, scores remain low and vary across July--November month slices. These results show that \bench supports both aggregate and stratified analyses, while avoiding stronger claims of monotonic temporal decay.

In summary, our contributions are as follows:
{\setlength{\leftmargini}{2.0em}
\begin{enumerate}[leftmargin=2.0em, itemsep=2pt, topsep=2pt, parsep=2pt, partopsep=2pt]
    \item  We release \textbf{\bench}, an expert-verified construction safety benchmark mined from 100{,}000+ raw image--text pairs, yielding 3,314 task instances with long-tail hazards, temporal/site metadata, and annotations covering hazard identification and hazard description.

    \item We provide a metadata\allowbreak-driven evaluation design that enables stratified temporal and site-level robustness analysis, together with executable evaluation scripts and a standardized LLM-as-a-judge scheme for scalable scoring of free-form outputs.

    \item We introduce \textbf{\method} as an automated selection pipeline that ranks informative candidates while enforcing diversity for expert verification, and we validate its effectiveness on a public dataset under data-budget constraints.
\end{enumerate}
}
\section{Related Work}

\subsection{Benchmarks for Multimodal LLMs}

\begin{figure*}[htbp]
    \centering
    \includegraphics[width=\linewidth]{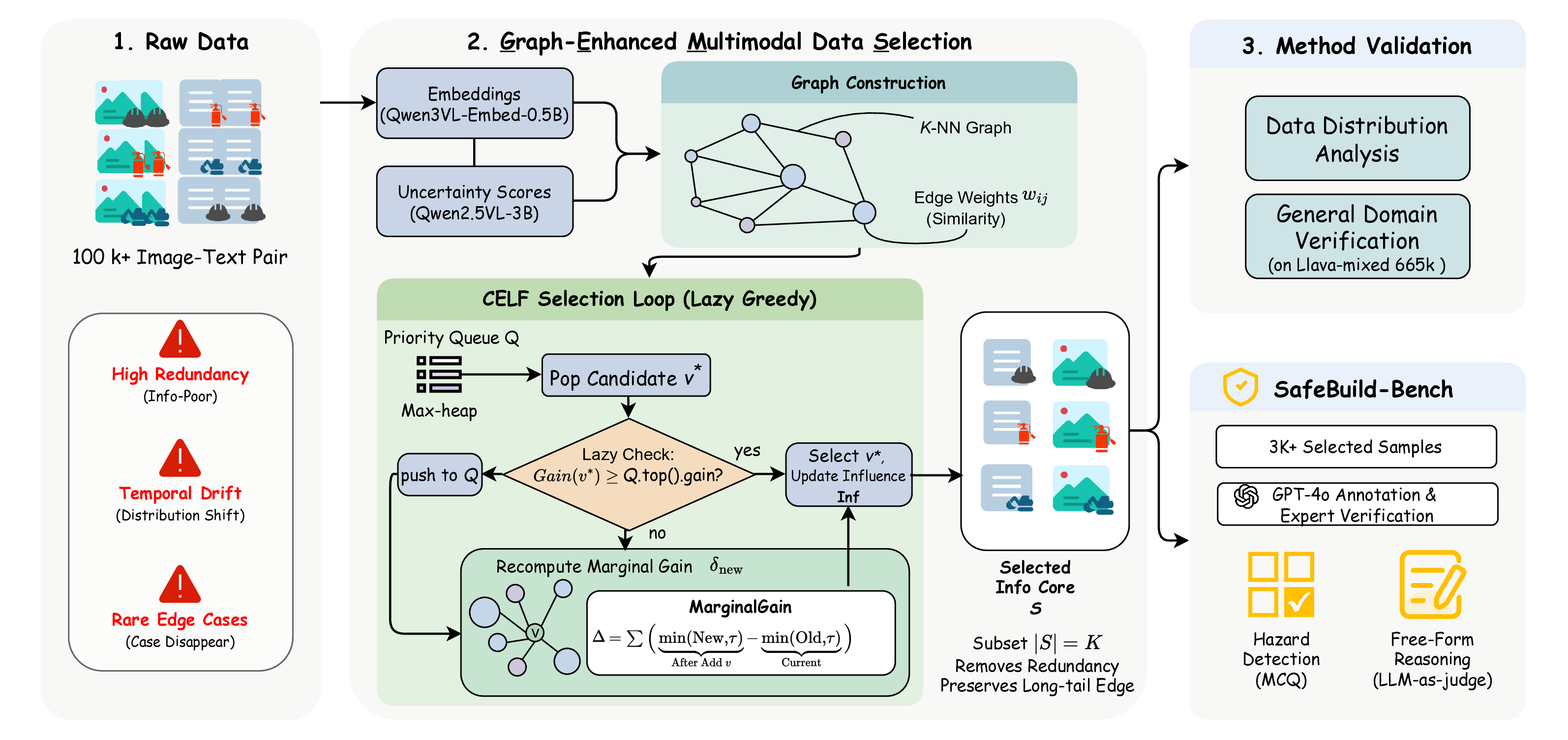}
    \caption{\textbf{Overview of the \method framework.} The pipeline involves: (1) \textbf{Graph Construction}: Modeling the data manifold using multimodal embeddings and uncertainty scores; (2) \textbf{Iterative Selection}: A CELF-based lazy greedy algorithm maximizes marginal gain $\Delta$, balancing uncertainty propagation with redundancy saturation ($\tau$); and (3) \textbf{Validation}: The method is verified on general dataset to ensure efficacy before constructing the domain-specific \bench.}
    \Description{A workflow diagram of GEMS showing graph construction from embeddings and uncertainty scores, iterative CELF-based subset selection, and validation before constructing SafeBuild-Bench.}
    \label{fig:pipeline}
\end{figure*}

Early vision--language evaluation largely centered on static VQA-style datasets~\cite{vqav2,gqa,vizwiz,textvqa}, 
and domain-focused benchmarks~\cite{rsvlm,scienceqa}, 
which primarily report answer-level accuracy under 
largely i.i.d.\ assumptions. 
As multimodal LLMs (MLLMs) emerged, broader and more diagnostic benchmarks were proposed. 
\textit{MME}~\cite{mme} and \textit{MMBench}~\cite{mmbench} aim to cover a wider range of perception and cognition 
skills with standardized protocols and controlled question design. 
In parallel, hallucination-oriented benchmarks such as \textit{POPE}~\cite{pope} and 
\textit{MMHal-Bench}~\cite{mmhalbench} evaluate object-level faithfulness and response reliability, 
increasingly leveraging \emph{LLM-as-a-judge} (or LLM-assisted normalization) to 
scale evaluation beyond exact-match metrics. 
Despite improved coverage and scalability, most existing benchmarks remain 
temporally static and do not explicitly stress-test MLLMs under realistic \emph{time} 
and \emph{site} shifts. \bench is designed to complement these efforts by 
focusing on temporal robustness and distribution shift in a high-stakes industrial domain.

\subsection{Data Selection for LLMs and MLLMs}
Data curation and selection have become essential for improving training efficiency and 
robustness in large-scale instruction tuning. Prior work spans influence-based and 
gradient-driven selection, uncertainty-/difficulty-aware sampling, and geometric coverage 
or coreset-style methods that reduce redundancy while preserving distributional diversity 
~\cite{less,coreset_survey,coincide,liu2024selectit}. A related line of hard-example and hard-negative mining improves discriminative training by emphasizing difficult samples, including online hard example mining, triplet-based hard negative selection, and focal reweighting for dense detection~\cite{shrivastava2016ohem,schroff2015facenet,lin2017focal}. \method is inspired by this focus on informative cases, but differs in target: it mines candidates for expert-verified benchmark construction rather than treating hard examples as labels for direct model optimization. Graph-based formulations have recently provided a way to model inter-sample dependence. In particular, the \emph{uncertainty-aware influence 
maximization} paradigm selects examples that are simultaneously informative 
(high uncertainty) and representative (strong influence on neighbors), enabling submodular 
objectives with greedy optimization and approximation guarantees~\cite{unimax_llm_doi}. 
For multimodal instruction data, recent methods~\cite{datatailor_arxiv,spice2026} explicitly balance 
informativeness, uniqueness, and representativeness to curate compact yet effective 
subsets for MLLM fine-tuning. Our pipeline, \method, follows this 
line but targets large, redundant industrial streams: it integrates epistemic uncertainty 
with a dual-graph structure to identify an informational core suitable for both efficient 
training and robust benchmarking.

\subsection{Hazard Understanding in Construction Sites}
Construction site monitoring has traditionally relied on standard computer vision tasks such as object detection for Personal Protective Equipment compliance and related hazard events~\cite{ppe_fall_detection}. While effective for closed-set categories, these supervised approaches typically require dense annotations, for example, bounding boxes, and scale poorly when new hazard types or safety rules must be added.
More recently, vision--language models have been explored for open-vocabulary or zero-shot safety assessment in construction settings~\cite{10.1145/3746252.3761652,safetogether,vlmsafe,rabbi2024ai}. However, existing benchmarks often lack a public, expert-verified evaluation set with explicit protocols for time and site shift, and they rarely emphasize long-tail, high-impact hazards within large and redundant inspection streams. \textbf{\bench} complements this line by providing an expert-verified public benchmark mined from real inspection archives, with hierarchical tasks covering hazard identification and hazard description, and with retained temporal metadata to support robustness analysis across time periods and sites.

\section{Construction Pipeline for \bench }\label{sec:methodology}

Creating a robust industrial benchmark requires moving beyond random sampling. We propose a systematic construction pipeline, as illustrated in Figure \ref{fig:pipeline}, comprising four stages: Raw Data Acquisition, Graph-Enhanced Mining (\method), Method Validation on General Domain dataset, and Expert-in-the-Loop Annotation, separated in the following subsections.

\subsection{Raw Data Acquisition and Preprocessing}
\label{sec:data_acquisition}
In collaboration with multiple large-scale construction sites, we curated an extensive dataset of image-text pairs over a five-month period from July to November 2025. These data were collected during the safety inspection phases of the construction process. Multiple safety experts and field personnel were involved in the acquisition and subsequent verification of the collected samples.
The raw dataset comprises over \textbf{100,000} image-text pairs sampled from more than 50 construction locations. This collection encompasses diverse scenes (e.g., scaffolding, foundation pits, tower cranes), various facilities (e.g., tower cranes, scaffolding, distribution boxes), and a range of environmental conditions (e.g., nighttime, rainy, and foggy weather).
To ensure strict adherence to privacy regulations, an automated anonymization pipeline was implemented. All identifiable faces and license plates were detected and obscured using a specialized obfuscation model. Furthermore, a random subset of the data was manually verified to ensure the complete absence of Personally Identifiable Information (PII).

\subsection{The \method Selection Engine} 
\label{sec:GEMS_method}
We propose \textbf{\method} (\underline{G}raph-\underline{E}nhanced \underline{M}ultimodal \underline{S}election), a data-centric framework distilling the ``informational core'' from redundant industrial streams. As shown in Figure \ref{fig:pipeline}, \method comprises three stages: (1) Multimodal Representation Learning, (2) Epistemic Uncertainty Quantification, and (3) Graph-Theoretic Submodular Selection.

\noindent \textbf{Multimodal Representation Learning.} 
To capture industrial scene semantics, we map image--text pairs into a unified dense vector space. Let $\mathcal{D} = \{(x_i, t_i)\}_{i=1}^N$ denote the raw pool, where $x_i$ is the image and $t_i$ is the paired inspection text or hazard description recorded for that image. We employ the \textsc{Qwen3-VL-Embedding}~\cite{qwen3vlembedding} encoder to extract fused embeddings. Unlike unimodal approaches that process vision and language separately, we utilize the model's native capability to obtain a joint representation:
\begin{equation}
\mathbf{z}_i = \mathcal{F}_{\theta}(x_i, t_i) \in \mathbb{R}^d,
\end{equation}
where $\mathcal{F}_{\theta}$ represents the encoder with frozen parameters. These high-dimensional embeddings ($\mathbf{z}_i$) serve as the geometric basis for our topological analysis.

\noindent \textbf{Epistemic Uncertainty Quantification.}
\label{sec:uncertainty} \method uses a proxy model (e.g., \textsc{Qwen2.5-VL-3B-Instruct}~\cite{qwen25_vl}) to obtain a \textbf{confusion signal} for candidate mining. Under a fixed hazard-analysis prompt $p$ (e.g., \textit{``Identify safety hazards''}), let $y$ be the generated response sequence of length $L$. We define the uncertainty score $u_i$ as the length-normalized negative log-likelihood (NLL): 
\begin{equation} 
    u_i = -\frac{1}{L} \sum_{t=1}^{L} \log P_{\phi}(y_t \mid y_{<t}, x_i, p).
\end{equation} 
A high $u_i$ means that the proxy assigns low probability to its own hazard-analysis response. We do not interpret this value as calibrated hazardness or as a final label: blur, rain, occlusion, or domain mismatch can also raise NLL. In \method, NLL only proposes potentially informative candidates, which are then filtered by graph diversity, safety relevance, visual clarity, ambiguity resolution, and final expert verification.

\noindent \textbf{Graph Construction and Optimization.}
We first construct a sparse $k$-Nearest Neighbor ($k$-NN) graph $G=(V, E)$ using the cosine similarity of embeddings $\mathbf{z}_i$, where weight $w_{ij}$ represents semantic proximity. For efficiency ($N > 100k$), we employ GPU-accelerated FAISS~\cite{faiss}. Our objective maximizes the total information covered by a subset $S \subseteq V$, where the "influence" received by node $j$ is defined as $\text{Inf}_j(S) = \sum_{i \in S} w_{ij} \cdot u_i$. To enforce diversity, we apply a saturation threshold $\tau$ to model \textit{diminishing returns}, yielding the global utility function:
\begin{equation}
    F(S) = \sum_{j \in V} \min \left( \text{Inf}_j(S), \tau \right).
\end{equation}
This monotone and submodular function guarantees a $(1 - 1/e)$-approximation via greedy strategy. Naive greedy selection would require $O(K \cdot N)$ marginal-gain evaluations. To scale \method, we employ the Cost-Effective Lazy Forward (CELF) algorithm~\cite{celf} (Algorithm \ref{alg:method}), which maintains a priority queue and uses submodularity-based lazy checks to reduce repeated recomputation in practice.

Overall, \method combines a proxy-model confusion signal with graph-theoretic selection to distill a compact yet informative subset from massive, redundant industrial streams. High-NLL nuisance cases, low-NLL but uninformative cases, and overconfident proxy outputs are not trusted automatically: mining only ranks candidates before human-facing quality control. This subset is therefore used as an expert-review queue rather than as an automatic benchmark labeler.

\subsection{\bench Construction}
\label{sec:safebuild-bench}
Building upon the curated construction safety data and the validated \method selection engine, we introduce \bench, a benchmark for evaluating vision-language models on realistic construction safety understanding tasks. Every instance in \bench retains its original collection timestamp and site identifier as metadata, so users can partition the evaluation set by month or site to measure robustness under different deployment slices. The complete benchmark, together with the metadata schema and executable evaluation scripts, is publicly released to facilitate reproducible and stratified robustness analysis.

\begin{figure}[tbp]
    \centering
    \includegraphics[width=.98\linewidth]{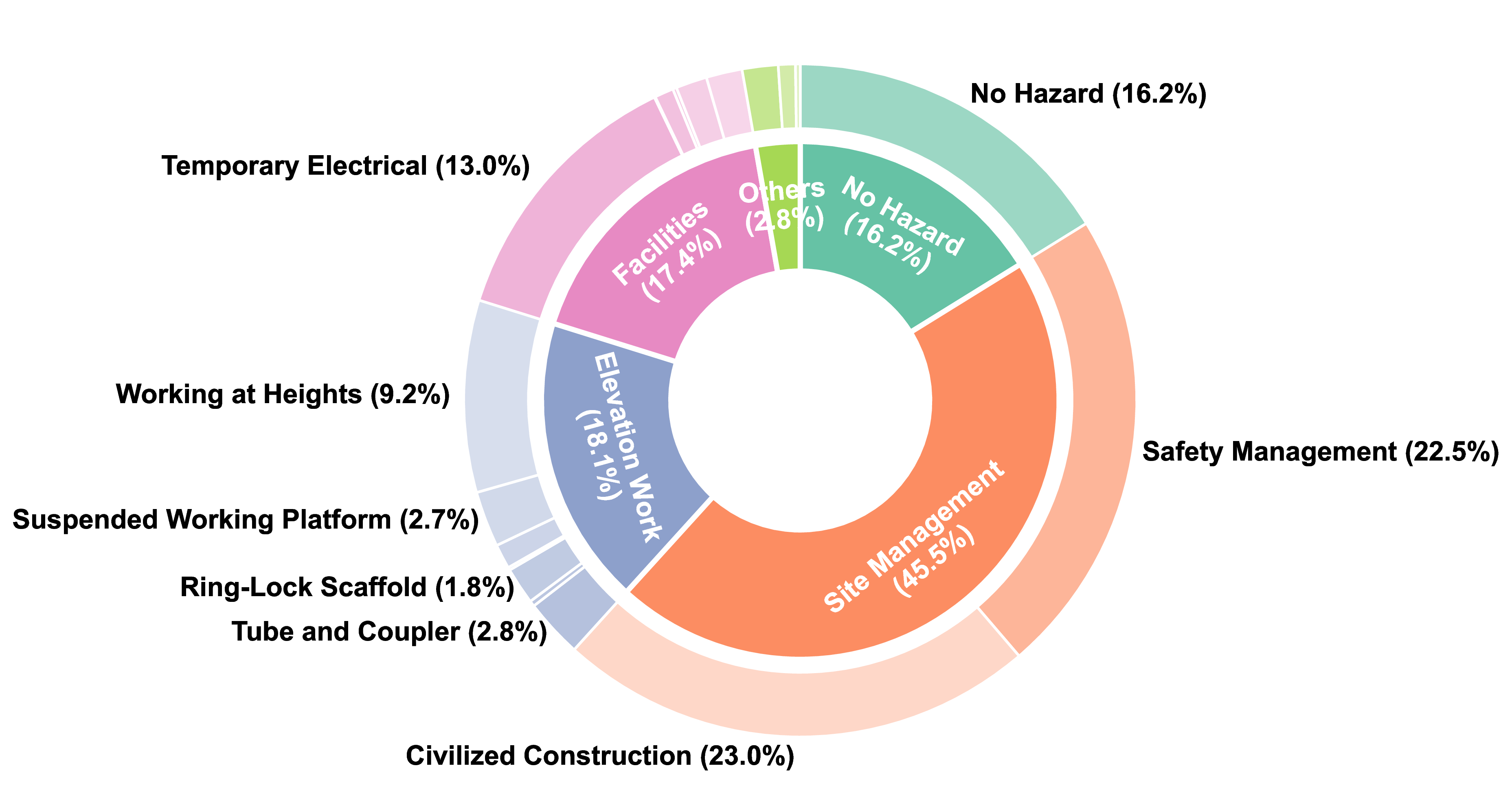}
    \caption{Category distribution of samples in \bench. The benchmark covers 19 fine-grained hazard categories, which are further grouped into five higher-level safety domains: Site Management, Elevation Work, Facilities, No Hazard, and Others.}
    \Description{A sunburst chart showing the distribution of SafeBuild-Bench samples across five high-level domains and nineteen fine-grained construction hazard categories.}
    \label{fig:category}
\end{figure}

\noindent\textbf{Sample Selection.} 
From the raw data pool, we select over 3,000 images to form \bench using the \method selection engine, yielding 3,314 task instances. The selection process prioritizes safety relevance, visual clarity, and coverage of diverse hazard types. Each selected sample is reviewed to ensure that the depicted scenario supports a grounded safety assessment and task formulation. As shown in Figure~\ref{fig:category}, the selected samples span 19 fine-grained hazard categories, organized into five higher-level safety domains, ensuring broad and structured coverage of construction safety scenarios.

\noindent\textbf{Annotation and Expert Verification.}
All hazard labels and reference descriptions in \bench are defined by construction safety experts. For each selected sample, experts assign the primary hazard category and provide a concise reference description following established construction safety inspection standards. Ambiguous cases are discussed and resolved through expert review to ensure consistency across categories and descriptions.

To reflect realistic inspection ambiguity, hazard categories are organized into expert-defined confusion groups. Each group consists of visually or procedurally similar hazard types that are frequently misidentified in practice, such as scaffold-related violations, temporary electrical hazards, site-management issues, and safe/no-hazard scenes. These groups are used to construct challenging multiple-choice instances by pairing the correct hazard category with visually and semantically similar alternatives, rather than arbitrary distractors.

\bench consists of two complementary tasks that evaluate both categorical recognition and descriptive understanding of construction safety hazards.

\noindent\textbf{Hazard Identification (MCQ).}  
This task evaluates a model’s ability to identify the primary construction safety hazard from visually and semantically similar alternatives. For each image, one correct hazard category is paired with three distractor categories sampled from the corresponding expert-defined confusion group. Models are required to select a single best answer. The task contains 2,200 instances and is evaluated using \textbf{Accuracy} and \textbf{Macro-Recall}. Accuracy measures the proportion of correctly identified hazards, while Macro-Recall computes the average recall across all hazard categories to assess robustness under class imbalance.

\noindent\textbf{Hazard Description (Free-form).}  
This task evaluates a model’s ability to generate precise and informative descriptions of construction safety hazards, or to correctly confirm the absence of hazards when applicable. The task contains 1,114 instances. Performance is evaluated using two metrics: \textbf{Hazard Detection Rate}, which measures whether the model correctly identifies the presence or absence of a safety hazard, and \textbf{Description Quality}, which assesses the semantic completeness and specificity of the generated description. Description Quality is scored on a five-point scale and linearly normalized to $[0,1]$. We use a fixed \textsc{GPT-4o} rubric-based judge that receives the model output, expert-written reference description, hazard-presence label, and scoring criteria. A 60-case two-human audit of \textsc{Kimi-K2.5} description outputs finds judge-human hazard agreement of 88.3\%--90.0\% and within-one agreement on description quality of 90.0\%--96.7\%, close to human-human agreement on the same sample. The judge prompt, rubric fields, audit summaries, and disagreement cases are released with the benchmark package.

\begin{algorithm}[t]
\caption{\method Selection via CELF (Lazy Greedy)}
\label{alg:method}
\begin{algorithmic}[1]
\Require Indices $V$, Embeddings $\mathcal{Z}$, Uncertainty $\mathcal{U}$, Budget $K$, Saturation $\tau$.
\Ensure Selected subset $S \subseteq V$ with $|S| = K$.

\State \textbf{Init:} Construct $k$-NN $G$, weights $w_{ij}$; $S \leftarrow \emptyset$, $\mathbf{Inf} \leftarrow \mathbf{0}_N$, $\mathcal{Q} \leftarrow \text{PQueue}()$.

\For{$v \in V$} \hfill \textcolor{gray}{\footnotesize // Step 3: Initial Pass}
    \State $\mathcal{Q}.\text{push}(\Call{MarginalGain}{v, \mathbf{Inf}}, v)$
\EndFor

\While{$|S| < K$} \hfill \textcolor{gray}{\footnotesize // Step 4: CELF Selection Loop}
    \State $v^*, \delta_{old} \leftarrow \mathcal{Q}.\text{pop}()$; \quad $\delta_{new} \leftarrow \Call{MarginalGain}{v^*, \mathbf{Inf}}$
    
    \If{$\delta_{new} \geq \mathcal{Q}.\text{top}().gain$} \hfill \textcolor{gray}{\footnotesize // Lazy property check}
        \State $S \leftarrow S \cup \{v^*\}$; \quad $\mathbf{Inf} \leftarrow \Call{UpdateInfluence}{v^*, \mathbf{Inf}}$
    \Else
        \State $\mathcal{Q}.\text{push}(\delta_{new}, v^*)$ \hfill \textcolor{gray}{\footnotesize // Re-insert with new gain}
    \EndIf
\EndWhile
\State \Return $S$

\Statex
\Function{MarginalGain}{$v, \mathbf{Inf}$}
    \State $\Delta \leftarrow 0$
    \For{$j \in \text{Neighbors}(v)$}
        \State $p \leftarrow w_{vj} \cdot u_v$; $\Delta \leftarrow \Delta + (\min(\mathbf{Inf}_j + p, \tau) - \min(\mathbf{Inf}_j, \tau))$
    \EndFor
    \State \Return $\Delta$
\EndFunction
\end{algorithmic}
\end{algorithm}

\vspace{-1.25em}
\section{Experiments}\label{sec:experiments}

\begin{table*}[t]
\centering
\caption{\textbf{Main Results on \bench.} 
We evaluate models on two core capabilities: \textit{Hazard Identification} and \textit{Hazard Description}. 
We report Global Accuracy and Macro-Recall for identification, and Hazard Detection Rate and Description Quality for description. Prices for closed-source models are reported as input/output costs per 1 million tokens (USD), based on official API pricing. The \textbf{Overall} score is the macro-average of the two tasks, reflecting the comprehensive safety capability.
The highest score in each column is shown in \textbf{bold}, and the second-highest score is \underline{underlined}.} 
\label{tab:safebuild_main}

\resizebox{\linewidth}{!}{
\setlength{\tabcolsep}{6pt}
\renewcommand{\arraystretch}{1.05}
\begin{tabular}{lcccccc}
\toprule
\multirow{2}{*}{\textbf{Model}} &
\multirow{2}{*}{\textbf{Cost(\$) / Params}} &
\multicolumn{2}{c}{\textbf{Hazard Identification}} &
\multicolumn{2}{c}{\textbf{Hazard Description}} &
\multirow{2}{*}{\textbf{Overall}} \\
\cmidrule(lr){3-4} \cmidrule(lr){5-6}
& &
\textbf{Accuracy} & \textbf{Macro-Recall}
& \textbf{Hazard Detection Rate} & \textbf{Description Quality}
& \\
\midrule

\rowcolor{proprietary}
\multicolumn{7}{c}{\textit{Closed-Source Models}} \\
\textsc{Gemini-3-Flash-Preview}~\cite{google2025gemini3flash}
& 0.5 / 3
& \textbf{55.8} & \underline{65.8} & 67.6 & 48.4 & \underline{59.4} \\
\textsc{Qwen3-VL-Plus}~\cite{qwen3_vl}
& 0.1 / 1.4
& 40.3 & 53.7 & \textbf{79.9} & \textbf{62.5} & 59.1 \\
\textsc{Claude-4.5-Sonnet}~\cite{anthropic2025claudesonnet45}
& 3.0 / 15.0
& 48.3 & 56.2 & 64.4 & 50.6 & 54.9 \\
\textsc{GPT-4o}~\cite{gpt4o}
& 2.5 / 10.0
& 34.3 & 43.8 & 66.5 & 50.7 & 48.8 \\
\midrule

\rowcolor{opensource}
\multicolumn{7}{c}{\textit{Open-Source Models}} \\
\textsc{Kimi-K2.5}~\cite{kimiteam2026kimik25visualagentic}
& 1T
& \underline{48.7} & \textbf{67.7} & \underline{72.6} & \underline{53.7} & \textbf{60.7} \\
\textsc{Qwen3-VL-235B-A22B-Instruct}~\cite{qwen3_vl}
& 235B
& 40.6 & 51.6 & 62.1 & 51.3 & 51.4 \\
\textsc{Qwen3-VL-8B-Instruct}~\cite{qwen3_vl}
& 8B
& 35.6 & 40.1 & 42.2 & 36.4 & 38.6 \\
\textsc{Qwen2.5-VL-7B-Instruct}~\cite{qwen25_vl}
& 7B
& 36.7 & 36.1 & 57.2 & 42.1 & 43.0 \\
\textsc{Qwen3-VL-4B-Instruct}~\cite{qwen3_vl}
& 4B
& 33.8 & 42.6 & 55.8 & 42.6 & 43.7 \\
\textsc{GLM-4.6V}~\cite{vteam2025glm45vglm41vthinkingversatilemultimodal}
& 106B
& 36.8 & 40.2 & 59.6 & 39.3 & 44.0 \\
\textsc{GLM-4.5V}~\cite{vteam2025glm45vglm41vthinkingversatilemultimodal}
& 106B
& 33.7 & 40.5 & 57.7 & 37.2 & 42.3 \\
\textsc{Gemma-3-12B-IT}~\cite{team2025gemma}
& 12B
& 27.7 & 40.0 & 55.8 & 40.5 & 41.0 \\
\textsc{InternVL3-8B}~\cite{internvl3}
& 8B
& 38.6 & 35.9 & 50.8 & 38.6 & 41.0 \\
\textsc{MiniCPM-V-4.5}~\cite{minicpm}
& 8B
& 33.9 & 43.5 & 45.9 & 31.7 & 38.7 \\
\textsc{LLaVA-1.5-7B-HF}~\cite{llava}
& 7B
& 25.3 & 29.8 & 38.2 & 27.2 & 30.1 \\
\bottomrule
\end{tabular}
}
\end{table*}

\subsection{Experimental Setup}\label{sec:setup}

We evaluate the proprietary and open-source vision-language models listed in Table~\ref{tab:safebuild_main}. All models are tested zero-shot on \bench, without training or fine-tuning on benchmark data. We use identical prompt templates and set decoding temperature to 0.2 when the provider allows it; \textsc{Kimi-K2.5} is evaluated at its supported temperature of 1.0, and other provider-specific parameters follow the official defaults. For hazard-description evaluation, we use the fixed \textsc{GPT-4o} rubric-based judge described in Section~\ref{sec:safebuild-bench}. Closed-source and open-source models are grouped separately in Table~\ref{tab:safebuild_main} for readability.

\subsection{\method Validation}

Before presenting the \bench results, we first validate the reliability of our construction pipeline. We hypothesize that if \method can effectively identify the ``informational core'' of a dataset, then a model trained on \method-selected data should exhibit superior robustness even with minimal samples. We evaluate our data selection method on two widely-used public instruction-tuning datasets with contrasting properties to stress-test the selection efficacy under different compression scenarios:

{\setlength{\leftmargini}{2.0em}
\begin{enumerate}[leftmargin=2.0em, itemsep=2pt, topsep=2pt, parsep=2pt, partopsep=2pt]
    \item \textbf{LLaVA-Instruct-150K} (Homogeneous Setting): This dataset serves as a test bed for \textit{in-domain homogeneous data compression}. It contains approximately 158K samples generated by GPT-4~\cite{gpt4_technical_report} based on the COCO~\cite{coco} dataset. The data is relatively homogeneous, primarily spanning conversation, detailed description, and complex reasoning tasks grounded in daily life images.
    
    \item \textbf{LLaVA-Mixed-665K} (Heterogeneous Setting): In contrast, this dataset represents a \textit{cross-task, cross-modality mixture compression} scenario. It is a heterogeneous collection (665K samples) combining LLaVA instructions with multiple academic datasets (e.g., VQAv2~\cite{vqav2}, OCR-VQA~\cite{ocr-vqa}, RefCOCO~\cite{refCOCO}). It also introduces task-specific response formatting prompts, adding to the distributional complexity.
\end{enumerate}}

We validate \method using LLaVA-v1.5-7B~\cite{llava} as the base. Comparisons include: (1) Random Selection (standard baseline); (2) DataTailor~\cite{datatailor_arxiv}; and (3) Full Data Training (ceiling). Evaluation metrics include robustness-oriented benchmarks: VizWiz~\cite{vizwiz} (real-world noise), MMMU~\cite{yue2024mmmu} (expert reasoning), MME~\cite{mme} (perception \& cognition), and POPE~\cite{pope} (hallucination). We group these metrics into \textit{General Perception}, \textit{Robustness \& Reliability}, and \textit{Knowledge \& Reasoning}. The overall score is the average across all benchmarks.

\noindent\textbf{Implementation Details.} For \method, we set the uncertainty weight to $\lambda=0.6$ and use \textsc{Qwen2.5-VL-3B} as the proxy model. We also ablate the uncertainty weight on LLaVA-Mixed-665K by testing $\lambda \in \{2, 4\}$, which increasingly up-weights model confusion in the selection objective. For all fine-tuned models, we freeze the vision tower and projector, and apply LoRA (rank 64, $\alpha$ 128) to all linear layers. Models are fine-tuned for 3 epochs with a learning rate of 2e-4, an effective batch size of 64, and a cosine schedule.

\begin{table*}[t]
\centering
\caption{\textbf{\method validation on public instruction-tuning datasets.} \textsc{LLaVA-v1.5-7B} is fine-tuned on subsets.
\textbf{MME} is the aggregate of perception and cognition scores.
\textbf{Avg.\ \%} denotes average relative performance versus the Full Data baseline.
Among data-efficient methods (excluding Full Data), the highest score in each column is shown in \textbf{bold} and the second-highest is \underline{underlined}.}
\label{tab:results_150k}\label{tab:results-665k}
\resizebox{\textwidth}{!}{%
\setlength{\tabcolsep}{7pt}
\renewcommand{\arraystretch}{1.18}
\begin{tabular}{l c rrr rr rrr r}
\toprule
\multirow{2}{*}{\textbf{Method}} & \multirow{2}{*}{\textbf{Data Size}} & \multicolumn{3}{c}{\textbf{General Perception}} & \multicolumn{2}{c}{\textbf{Robustness \& Reliability}} & \multicolumn{3}{c}{\textbf{Knowledge \& Reasoning}} & \multirow{2}{*}{\textbf{Avg.\ \%}} \\
\cmidrule(lr){3-5} \cmidrule(lr){6-7} \cmidrule(lr){8-10}
 & & \textbf{VQAv2} & \textbf{GQA} & \textbf{MME} & \textbf{VizWiz} & \textbf{POPE} & \textbf{MMMU} & \textbf{SciQA} & \textbf{TextVQA} & \\
\midrule
\rowcolor{baselinegray}
\multicolumn{11}{l}{\textit{\textbf{(a) LLaVA-Instruct-150K\quad (Homogeneous Setting)}}} \\
\midrule
Full Data          & 100\% (158k) & 73.7 & 60.6 & 1780.8 & 54.1 & 85.5 & 33.7 & 66.3 & 47.4 & 100.0 \\
Random Selection   & 20\% (32k)   & \textbf{75.6} & \underline{60.5} & 1675.0 & 42.3 & 85.7 & 32.8 & \underline{65.8} & 47.6 & 97.5 \\
\rowcolor{oursgreen}
\multicolumn{11}{l}{\quad\textit{\textbf{Ours: \method}}} \\
\textbf{\method}   & 1\% (1.5k)   & 75.3 & 60.4 & 1699.8 & 52.1 & \textbf{86.5} & 34.2 & \textbf{66.2} & \underline{47.7} & 99.5 \\
\textbf{\method}   & 10\% (15k)   & \underline{75.4} & 60.3 & \textbf{1768.8} & \underline{53.0} & \underline{86.1} & \underline{35.3} & 65.3 & 47.1 & \underline{100.2} \\
\textbf{\method}   & 20\% (32k)   & \textbf{75.6} & \textbf{60.6} & \underline{1755.2} & \textbf{55.2} & 85.8 & \textbf{35.7} & \underline{65.8} & \textbf{47.8} & \textbf{101.1} \\
\midrule
\rowcolor{baselinegray}
\multicolumn{11}{l}{\textit{\textbf{(b) LLaVA-Mixed-665K\quad (Heterogeneous Setting)}}} \\
\midrule
Full Data                          & 100\% (665k) & 79.1 & 63.0 & 1744.8 & 47.8 & 86.4 & 32.8 & 70.0 & 58.2 & 100.0 \\
Random Selection                   & 8\% (50k)    & 73.7 & 55.0 & 1675.0 & 42.3 & 85.7 & 32.2 & \underline{70.0} & \textbf{53.1} & 93.3 \\
DataTailor~\cite{datatailor_arxiv} & 8\% (50k)    & \underline{75.0} & 57.7 & \textbf{1823.7} & 46.3 & 82.1 & 33.9 & \textbf{70.9} & \textbf{53.1} & 96.9 \\
\rowcolor{ablationtan}
\multicolumn{11}{l}{\quad\textit{\textbf{Ablation: Uncertainty Weight}}} \\
\method w/$\lambda\!=\!2$          & 8\% (50k)    & 74.9 & \underline{60.4} & 1766.6 & 49.4 & \underline{86.0} & 34.8 & 64.8 & 47.5 & 97.4 \\
\method w/$\lambda\!=\!2$          & 1\% (6k)     & \textbf{75.1} & \textbf{60.8} & 1691.8 & 49.3 & 85.9 & 34.2 & 64.7 & \underline{47.8} & 96.9 \\
\method w/$\lambda\!=\!4$          & 1\% (6k)     & 72.3 & 58.4 & 1642.7 & 52.9 & 84.7 & 34.9 & 64.3 & 45.9 & 96.0 \\
\rowcolor{oursgreen}
\multicolumn{11}{l}{\quad\textit{\textbf{Ours: \method}}} \\
\textbf{\method}                   & 1\% (6k)     & \textbf{75.1} & \underline{60.4} & \underline{1805.0} & \underline{53.7} & \textbf{86.2} & 34.8 & 65.5 & 47.3 & \textbf{98.3} \\
\textbf{\method}                   & 8\% (50k)    & 74.8 & 60.1 & 1781.9 & 51.6 & 85.9 & \underline{36.0} & 65.4 & 46.2 & 97.8 \\
\textbf{\method}                   & 15\% (100k)  & 74.8 & 60.3 & 1728.3 & \textbf{54.0} & 85.3 & \textbf{36.1} & 66.3 & 46.7 & \underline{98.1} \\
\bottomrule
\end{tabular}%
}
\end{table*}

\noindent\textbf{Results on Homogeneous Data (LLaVA-Instruct-150K).} 
Table \ref{tab:results_150k}(a) demonstrates the efficacy of \method under a relatively homogeneous instruction distribution. 
First, we observe an extreme \textit{data efficiency gain}: training with merely \textbf{1\%} of the data (1.5k samples) surpasses the \textbf{Random Selection} baseline using 20\% data (32k samples) on aggregate metrics (MME: 1699.8 vs. 1675.0). 
Notably, the 1\% subset achieves the highest \textbf{POPE} score (\textbf{86.5}) among all settings, including Full Data (85.5), suggesting that filtering out redundant captions significantly mitigates hallucination. 
When scaling to \textbf{20\%}, \method consistently breaks the performance ceiling of the full dataset, particularly on reasoning-intensive and noisy benchmarks, achieving \textbf{55.2} on VizWiz (+1.1 over Full) and \textbf{35.7} on MMMU (+2.0 over Full). 
Remarkably, \method achieves \textbf{99.5\%} of the full-data performance using only \textbf{1\%} of the samples (1.5k). At the 20\% scale, it further surpasses the full dataset with a relative score of \textbf{101.1\%}, suggesting that \method effectively filters out redundant or noisy data that hinders optimization.

\noindent\textbf{Results on Heterogeneous Mixture (LLaVA-Mixed-665K).}
Table \ref{tab:results-665k}(b) reports selection performance on the large-scale, multi-task LLaVA-665K dataset.
Under this highly redundant mixture, \method improves data efficiency on robustness-oriented benchmarks.
A model fine-tuned on 1\% of \method-selected data (about 6.6k samples) outperforms the full-data model (665k) on in-the-wild evaluation, for example, 53.7 on VizWiz (+5.9 over full data) and 34.8 on MMMU (+2.0 over full data).
Compared with the Random-8\% baseline, \method-1\% also yields a clear gain on MME (1805.0 vs.\ 1675.0).
Overall, \method retains 98.3\% of the full-data average while using only 1\% of the training samples.
These results suggest that selection can reduce redundancy in heterogeneous instruction mixtures while preserving the hard and informative content that matters for robustness evaluation.
We use this behavior as a proxy validation signal for constructing \bench: the same selection mechanism can prioritize informative and non-duplicate candidates in large industrial streams before expert verification.

\begin{table}[t]
\centering
\caption{\textbf{In-domain subset composition on a cached November mining pool.} At a matched 10\% budget from 8.2K candidates, full \method reduces redundancy and improves regulation diversity relative to uncertainty-only mining.}
\label{tab:gems_indomain}
\small
\setlength{\tabcolsep}{4pt}
\begin{tabular}{lccc}
\toprule
\textbf{Method} & \textbf{NN cosine$\downarrow$} & \textbf{Regs$\uparrow$} & \textbf{Top-3 share$\downarrow$} \\
\midrule
Uncertainty-only & 0.926 & 80 & 0.666 \\
\method & 0.825 & 305 & 0.559 \\
\bottomrule
\end{tabular}
\end{table}

We also compare against 100 matched-size random draws on the same cached pool: \method has lower within-subset redundancy (0.825 vs.\ $0.887\pm0.003$) and slightly better category balance (top-3 share 0.559 vs.\ $0.592\pm0.015$), while regulation diversity remains competitive (305 vs.\ $304\pm7$). This is subset-composition evidence for benchmark construction, not a full diversity-only ablation or end-to-end retraining study.

\subsection{Main Results}\label{sec:results}

Table~\ref{tab:safebuild_main} reports the main results on \bench across two core capabilities: Hazard Identification and Hazard Description. Overall, current vision-language models remain far from robust construction safety understanding. Even the best-performing models achieve Overall scores around 60, indicating substantial room for improvement across both identification and description tasks.

For \textbf{Hazard Identification}, performance varies notably across models. \textsc{Gemini-3-Flash-Preview} achieves the highest accuracy (55.8), while \textsc{Kimi-K2.5} attains the best Macro-Recall (67.7), suggesting stronger robustness under class imbalance. In contrast, several models exhibit a large gap between Accuracy and Macro-Recall, indicating that correct predictions are often concentrated in frequent categories while rare hazards remain challenging.

For \textbf{Hazard Description}, \textsc{Qwen3-VL-Plus} achieves the highest Hazard Detection Rate (79.9) and Description Quality (62.5). This suggests that strong descriptive ability does not necessarily correlate with categorical identification accuracy, as \textsc{Qwen3-VL-Plus} lags behind top models on identification metrics. Across models, Hazard Detection Rate is generally higher than Description Quality, indicating that models can often detect the existence of hazards but struggle to provide precise and informative descriptions.

Comparing closed-source and open-source models, we observe competitive performance from large open-source systems. \textsc{Kimi-K2.5} achieves the highest Overall score (60.7) among all evaluated models, outperforming several proprietary counterparts. However, smaller open-source models underperform, highlighting the importance of model scale for construction safety understanding.This trend is consistent across open-source models, where performance improvements are primarily driven by increases in model scale rather than architectural variations.

Taken together, these results reveal that construction safety remains a challenging domain for vision-language models. Strong performance on one task does not guarantee robustness on the other, underscoring the need for holistic evaluation across both hazard identification and hazard description capabilities.

\begin{table}[t]
\centering
\caption{\textbf{Month-stratified MCQ accuracy on \bench.} Fixed models show material July--November variation; this supports temporal slicing using released metadata, but is not a held-out-site or forward-chaining protocol.}
\label{tab:monthly_m}
\small
\setlength{\tabcolsep}{4pt}
\begin{tabular}{lccccc}
\toprule
\textbf{Model} & \textbf{Jul} & \textbf{Aug} & \textbf{Sep} & \textbf{Oct} & \textbf{Nov} \\
\midrule
\textsc{Gemini-3-Flash-Preview} & 52.2 & 60.1 & 62.4 & 60.6 & 55.7 \\
\textsc{Kimi-K2.5} & 45.6 & 51.3 & 53.9 & 50.0 & 49.8 \\
\bottomrule
\end{tabular}
\end{table}

The same month-specific pattern remains on a category-controlled macro metric over stable categories. Thus, temporal metadata in \bench is operational for quantitative stratified analysis, while more demanding protocols such as strict forward-chaining or held-out-site evaluation remain future extensions.
Across the broader model set, month-level accuracy ranges are typically about 8--13 points. For example, \textsc{Claude-Sonnet-4.5} varies from 46.4\% in July to 54.6\% in October and 47.3\% in November, \textsc{Qwen3-VL-Plus} ranges from 33.3\% to 45.9\%, and \textsc{GPT-4o} ranges from 27.3\% to 40.0\%. We interpret October cautiously because it contains only 66 MCQ items, but the pattern persists after restricting the analysis to stable categories. This supports reporting temporal slices alongside the aggregate score, since the same model can look materially different under different collection months.

\subsection{Error Analysis}\label{sec:analysis}
\begin{figure}[!t]
    \centering
    \includegraphics[width=.98\linewidth]{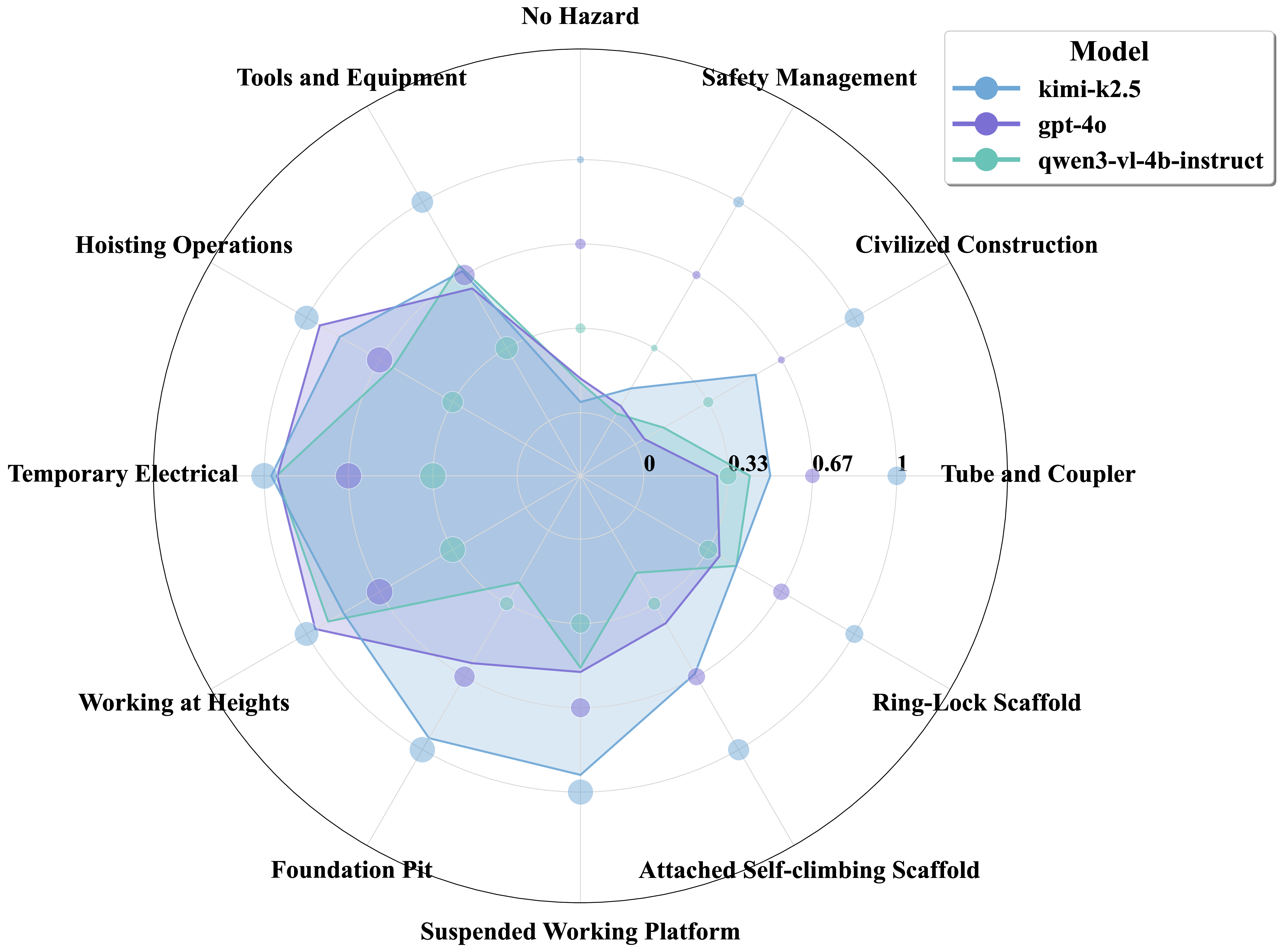}
    \caption{Category-wise identification accuracy on the \textit{Hazard Identification} task. The radar chart compares three representative models, \textsc{Kimi-K2.5}, \textsc{GPT-4o}, and \textsc{Qwen3-VL-4B-Instruct}, across major hazard categories. Each axis corresponds to a hazard category, and values indicate identification accuracy within that category.}
    \Description{A radar chart comparing hazard-identification accuracy by category for Kimi-K2.5, GPT-4o, and Qwen3-VL-4B-Instruct.}
    \label{fig:identification}
    \vspace{-1em}
\end{figure}

\textbf{Hazard Identification.}
Figure~\ref{fig:identification} illustrates category-wise identification accuracy for representative models on the Hazard Identification task. A clear pattern emerges: model performance varies substantially across hazard categories, indicating that identification errors are strongly category-dependent rather than uniformly distributed. We also observe consistent trends across different models, suggesting shared challenges inherent to specific hazard types.

Models generally perform better on hazard categories with distinctive visual cues and well-defined objects, such as \textit{Temporary Electrical} and \textit{Hoisting Operations}. In contrast, categories involving subtle structural differences or contextual safety rules, such as \textit{Tube and Coupler}, \textit{Ring-Lock Scaffold}, and \textit{Safety Management}, consistently exhibit lower accuracy across models. These categories often require fine-grained discrimination between visually similar configurations or rely on implicit safety norms, which remain challenging for current vision-language models.

Another notable failure mode is misclassifying \textit{No Hazard} cases. Despite the absence of explicit safety violations, models frequently over-predict hazards, suggesting a bias toward hazard presence. This tendency indicates limited calibration in distinguishing compliant construction scenarios from genuinely unsafe ones.

Comparing models, larger models such as \textsc{Kimi-K2.5} and \textsc{GPT-4o} show relatively more stable performance across categories, while smaller models like \textsc{Qwen3-VL-4B-Instruct} exhibit sharper performance drops in complex or ambiguous categories. This observation suggests that model capacity plays an important role in handling fine-grained hazard distinctions, although even large models remain far from reliable across all categories.

To check whether low MCQ scores mainly reflect annotation ambiguity, we audited 60 hard cases drawn from \textsc{Gemini-3-Flash-Preview} errors, prioritizing cases also missed by other strong models. In this stress-tested subset, 68.3\% of cases were still judged clear and 78.3\% kept the current gold label; even among 30 cases missed by all four reference models, 73.3\% were clear and 70.0\% kept the gold label. These results indicate that ambiguity exists, but current model weakness cannot be explained by widespread benchmark ambiguity alone.

\textbf{Hazard Description.}
We further analyze the Hazard Description task using \textsc{GPT-4o} as an automatic judge, which outputs a hazard-detection signal (HDR) and a normalized description-quality score. We define \emph{major} description failures as cases with $\mathrm{HDR}=0$ or $\mathrm{Final}\le 0.5$ (excluding judge errors and dataset-conflict cases). Under this criterion, \textsc{Kimi-K2.5} exhibits a 27.3\% major-failure rate (304/1114), while the smaller model \textsc{Qwen3-VL-4B-Instruct} rises to 44.4\% (495/1114), indicating a capacity gap in hazard narration.

Across both models, major failures are overwhelmingly driven by breakdowns at the hazard detection stage rather than deficiencies in linguistic expression. Almost all major failures coincide with $\mathrm{HDR}=0$ (303/304 for \textsc{Kimi-K2.5}; 493/495 for \textsc{Qwen3-VL-4B-Instruct}), suggesting that once a hazard is correctly detected, models usually produce descriptions of acceptable quality.

The two models exhibit distinct failure patterns. \textsc{Kimi-K2.5} often fails by describing hazards that appear plausible but do not satisfy the ground-truth safety criteria (52.0\% of major failures). False alarms in genuinely safe scenes account for a further 30.9\%. In contrast, \textsc{Qwen3-VL-4B-Instruct} is dominated by missed hazards, frequently responding with generic “no hazard” or “safe” statements when safety violations are present (70.7\%). The remaining cases mainly involve mismatched hazard descriptions (21.0\%).

For both models, failures are concentrated in scenarios that require contextual grounding rather than the recognition of a salient object, such as signage and housekeeping compliance, barriers and guardrails, scaffolding, and temporary electrical setups. These results indicate that hazard description remains challenging when safety violations are defined by implicit rules or subtle visual cues.

\vspace{-1em}

\section{Discussion and Limitations}
\label{sec:discussion}

Our results highlight a practical issue in construction safety monitoring: large inspection archives are often dominated by redundancy, while rare hazards that matter for deployment form a long tail. \bench is designed to make this setting measurable by retaining temporal and site metadata and enabling stratified analysis, which better reflects real deployment shift than a single aggregate score. The proxy studies and in-domain composition analysis suggest that selection can preserve robustness-relevant content under small data budgets, supporting a data-centric approach to benchmark construction.

\noindent\textbf{Additional audits.}
The added audit results bound this interpretation. Month-level MCQ slicing shows material July--November variation under fixed models, confirming that the released metadata supports quantitative stratified analysis rather than serving only as descriptive fields. The two-human description audit shows that the \textsc{GPT-4o} judge is directionally reliable: judge-human hazard agreement reaches 90.0\%, close to 88.3\% human-human agreement, and description-quality scores are within one point of the human reference in 90.0\% of cases. The hard-MCQ audit further suggests that low identification scores are not explained by pervasive annotation noise, since most stress-tested errors remain clear and keep the gold label. Finally, the November cached-pool composition check shows that \method is not merely uncertainty-only ranking: it lowers redundancy while preserving broad regulation coverage at a matched budget.

\noindent\textbf{Release scope.}
We separate the stable benchmark from supporting analysis packages. The former defines the evaluation artifact, while the latter documents the month-slicing, judge-audit, ambiguity-audit, and subset-composition evidence used to calibrate our claims. This organization keeps the added results reproducible without overstating them as complete forward-chaining, held-out-site, or multi-proxy sensitivity studies.

\noindent\textbf{Recommended reporting.}
For practical use, \bench should be reported with both aggregate and stratified scores. The aggregate score gives a compact comparison across systems, but the month, site, category, and task-level views are needed to diagnose deployment risk. In particular, the distinction between missed hazards and false alarms is important for construction safety: missed hazards indicate direct safety risk, while false alarms can make automated inspection systems difficult to trust in routine use. We therefore recommend that future evaluations report hazard-identification accuracy together with macro-recall, no-hazard calibration behavior, and the hazard-detection and description-quality components of the free-form task. This makes model comparisons more informative than a single leaderboard number and keeps the benchmark aligned with operational safety decisions.

Limitations remain. First, \bench contains 3,314 task instances from over 3,000 expert-verified images, and scaling will require sustained expert effort; our planned expansion is a rolling mine-then-verify protocol over new months and sites. Second, \method depends on proxy models for confusion scoring, and weak proxies may under-select certain hazard types; the current analysis does not include a full multi-proxy sensitivity study. Third, \bench is grounded in Chinese construction regulations. Many visual hazards, such as missing helmets, unstable scaffolds, fall risk, and suspended-load exposure, are physically cross-context, but transferring the benchmark recipe to other countries requires new taxonomies, expert guidelines, and validation.

\section{Conclusion}

We release \bench, a construction safety benchmark for evaluating MLLMs under realistic time and site variation. \bench is mined from large-scale inspection archives and curated into 3,314 task instances from over 3,000 expert-verified images, covering hazard identification and hazard description. Each instance retains temporal and site metadata, enabling stratified analysis with executable evaluation scripts and a standardized LLM-as-a-judge protocol.
To make construction feasible from redundant streams, we develop \textbf{\method}, a graph-enhanced mining pipeline that prioritizes informative and non-duplicate candidates for expert verification. In proxy studies on public instruction-tuning data, fine-tuning on a small \method-selected subset can match or exceed full-data training on robustness-oriented benchmarks, suggesting that selection can reduce redundancy while preserving hard cases. We also release the dataset, benchmark, curation codebase, and the evaluation scripts to support reproducible research on safety-focused multimodal models.

The release package also includes supporting analyses for temporal slicing, judge reliability, ambiguity, and subset composition, making the added evidence and its scope auditable.

\begin{acks}
This work was supported in part by the National Key R\&D Program of China (Grant No. 2023YFF0725001), and in part by the National Natural Science Foundation of China (Grant No. 92370204). We gratefully acknowledge Jianhua Yang (\texttt{zhbgs@gdzgy.com}) from Guangdong Zhonggong Architectural Design Institute Co., Ltd., Jian Lin (\texttt{gdzgjl@gdzgjl.com}) from Guangdong Zhonggong Project Management Co., Ltd., and Yan Liu (\texttt{gdzgjl@gdzgjl.com}) from Guangdong Zhonggong Project Management Co., Ltd. for their contributions to data collection and data verification in this project.
\end{acks}

\clearpage

\bibliographystyle{ACM-Reference-Format}
\balance
\bibliography{ref}

\appendix

\section{Release, Audit, and Ethics Notes}
\label{appendix:release_audit}

\noindent\textbf{Dataset composition and release.}
\bench contains over 3,000 expert-verified images and 3,314 task instances, including 2,200 multiple-choice hazard-identification items and 1,114 free-form hazard-description items. Each released instance includes the image, hazard category, expert-written reference, task metadata, temporal metadata, and site identifier needed for month-level or site-level stratification. The public package includes the benchmark schema, evaluation scripts, GEMS mining code, and supporting analysis packages used in the camera-ready revision.

\noindent\textbf{Supporting analyses.}
The release package separates benchmark data from rebuttal-time supporting analyses. The P0 temporal-slicing package contains the July--November 2025 month-level MCQ protocol and fixed-model scores; it is evidence for measurable temporal heterogeneity, not a forward-chaining or held-out-site experiment. The P1 package contains the judge prompt, rubric, reference fields, 60-case two-human audit summary, and representative disagreement cases. The P2 package contains the hard-MCQ ambiguity audit summary. The P3 package contains the November-only 8.2K cached-pool, 10\%-budget GEMS composition analysis and its uncertainty-only and random baselines.

\noindent\textbf{Ethics and intended use.}
Images were collected from construction inspection records and anonymized before release when personally identifying details such as faces or license plates were visible. \bench is intended for evaluating and improving safety-focused multimodal models. It should not be used for biometric identification, worker surveillance, punitive monitoring, or automated disciplinary decisions. Because the benchmark is grounded in Chinese construction regulations, deployment in other regulatory contexts requires fresh taxonomies, expert guidelines, and validation.

\section{Disclosure of AI Use}
In the preparation of this manuscript, we utilized AI tools to assist with linguistic polishing and the formatting of tables to improve readability. The authors have reviewed and edited the content, and take full responsibility for the accuracy and integrity of the published work.

\section{Benchmark Case Study}
\label{appendix:case_study}

\begin{figure*}[!t]
    \centering
    \tcbset{colback=white, colframe=gray!50!black, fonttitle=\bfseries, arc=2mm,
        boxsep=1.5pt, left=1.2mm, right=1.2mm, top=1.2mm, bottom=1.2mm}
    \begin{minipage}[t]{0.48\linewidth}
        \includegraphics[width=\linewidth,height=4.9cm]{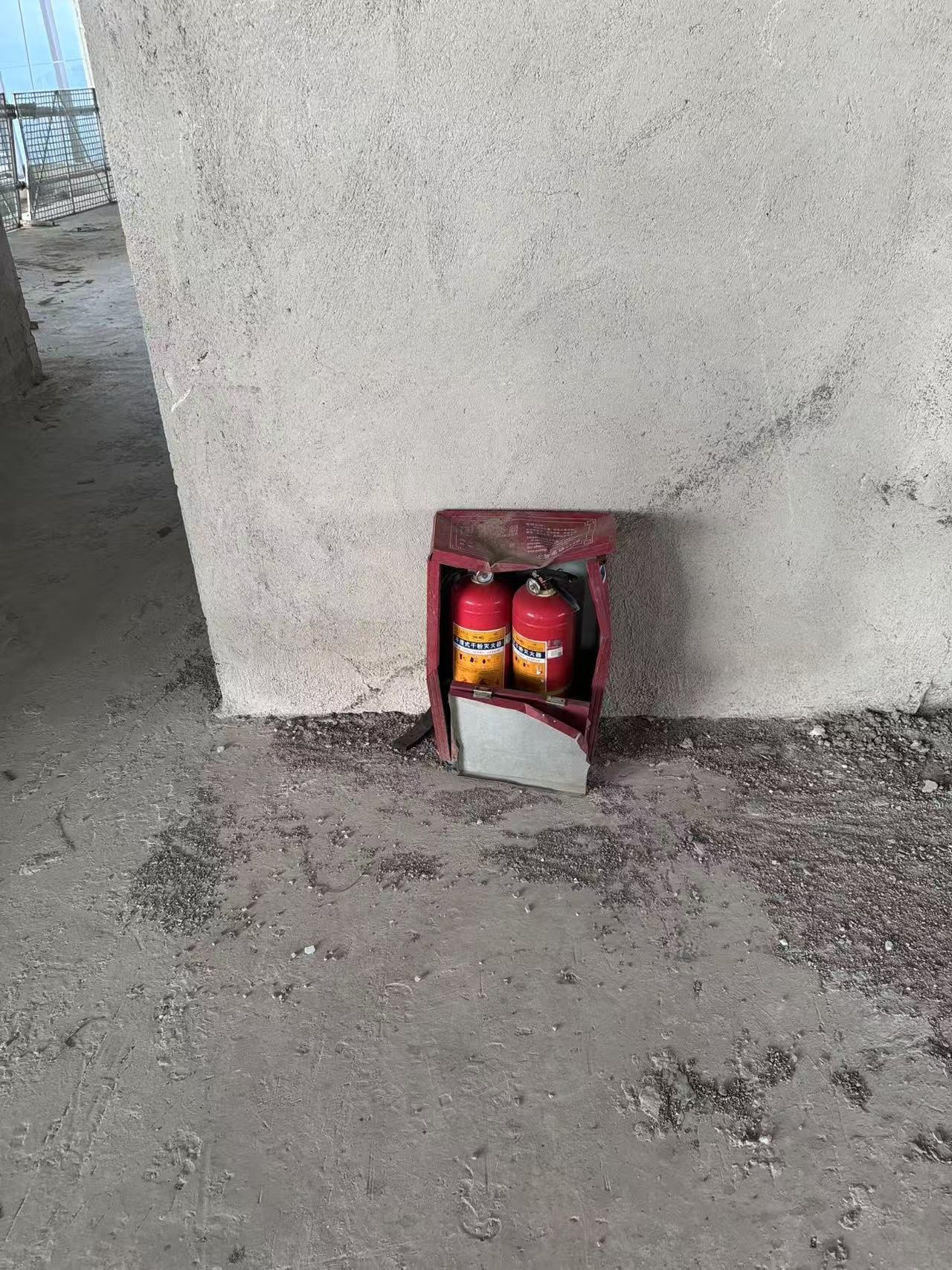}
        \vspace{-2mm}
        \begin{tcolorbox}[title=Task: Hazard Detection (MCQ), colbacktitle=blue!60!black]
            \footnotesize
            \textbf{Question:} Identify the primary construction safety hazard type shown in the image.
            \vspace{0.5mm}
            \textbf{Options:}
            \begin{itemize}[leftmargin=1.4em,itemsep=0pt,topsep=0pt]
                \item[A.] Temporary Electrical Supply
                \item[B.] \colorbox{green!20}{\textbf{No Hazard}} \textcolor{green!60!black}{\safebuildcheck}
                \item[C.] Safety Management
                \item[D.] Civilized Construction
            \end{itemize}
            \vspace{0.5mm}
            \textit{\textbf{Analysis:} Option B is correct because the construction site is orderly, with fire extinguishers properly stored and accessible. Safety measures appear to be in place.}
        \end{tcolorbox}
    \end{minipage}
    \hfill
    \begin{minipage}[t]{0.48\linewidth}
        \includegraphics[width=\linewidth,height=4.9cm]{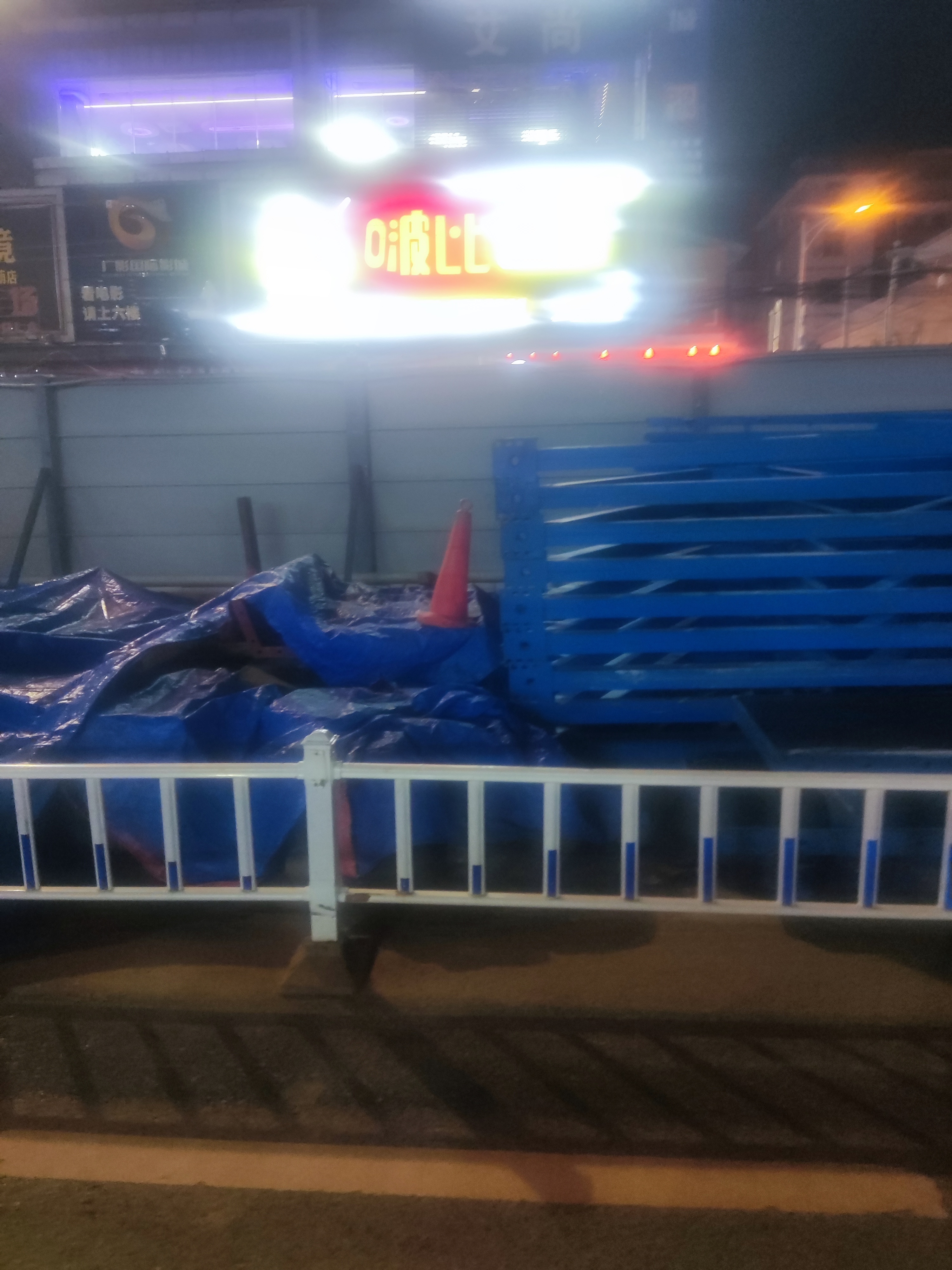}
        \vspace{-2mm}
        \begin{tcolorbox}[title=Task: Hazard Detection (MCQ), colbacktitle=blue!60!black]
            \footnotesize
            \textbf{Question:} Identify the primary construction safety hazard type shown in the image.
            \vspace{0.5mm}
            \textbf{Options:}
            \begin{itemize}[leftmargin=1.4em,itemsep=0pt,topsep=0pt]
                \item[A.] No Hazard
                \item[B.] Construction Tools and Equipment
                \item[C.] \colorbox{green!20}{\textbf{Safety Management}} \textcolor{green!60!black}{\safebuildcheck}
                \item[D.] Civilized Construction
            \end{itemize}
            \vspace{0.5mm}
            \textit{\textbf{Analysis:} Option C is correct because the area lacks proper safety signage to warn of potential hazards, and the equipment is covered with a tarp, which may conceal tripping hazards.}
        \end{tcolorbox}
    \end{minipage}

    \caption{\textbf{Qualitative Examples from \bench.} Two MCQ samples testing precise hazard identification. The correct option is highlighted.}
    \Description{Two construction-site multiple-choice examples with images, answer options, highlighted correct labels, and short safety explanations.}
    \label{fig:fig_mcq_sample}
\end{figure*}

\begin{figure*}[!t]
    \centering
    \tcbset{colback=white, colframe=gray!50!black, fonttitle=\bfseries, arc=2mm,
        boxsep=1.5pt, left=1.2mm, right=1.2mm, top=1.2mm, bottom=1.2mm}
    \begin{minipage}[t]{0.48\linewidth}
        \includegraphics[width=\linewidth,height=4.9cm]{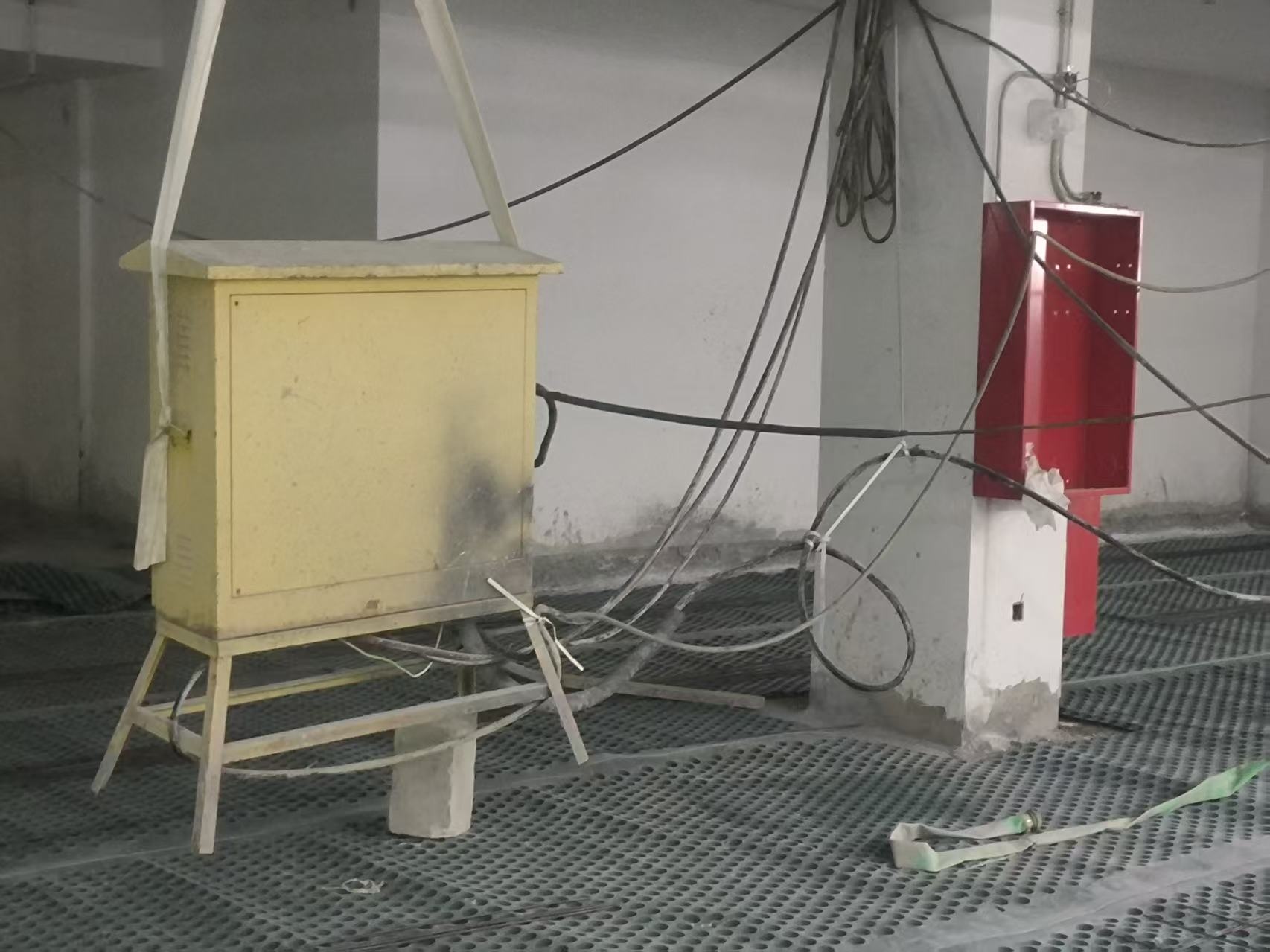}
        \vspace{-2mm}
        \begin{tcolorbox}[title=Task: Open-Ended Description \& Evaluation, colbacktitle=purple!70!black]
            \footnotesize
            \textbf{Ground Truth:} \textit{The \textbf{safety netting} is improperly installed, with \textbf{visible gaps} that could lead to falls from height.}
            \vspace{0.5mm}
            \hrule
            \vspace{0.5mm}
            \textbf{Model Prediction:}
            There is a hazard due to exposed and tangled \textbf{electrical cables}, which pose a risk of tripping, electrical shock, or fire. The suspended \textbf{yellow box} appears unstable, increasing the risk of falling and injury.\textcolor{green!60!black}{\safebuildcheck}
            \par\vspace{1mm}
            \textbf{\textsc{LLM-as-a-judge Output:}}\par
            \begin{description}[leftmargin=0.1em,itemsep=0pt,topsep=0pt]
                \item[Hazard Detection:] \textbf{1}
                \item[Quality Score:] \textbf{4/5} \textcolor{gray}{(Miss Key Words)}
                \item[Final Score:] \colorbox{green!10}{\textbf{0.875}}
            \end{description}
            \textit{\textbf{Reasoning:} The predicted description correctly identifies the existence of hazards, including tangled cables and the instability of the suspended box. It explains risks such as tripping, electrical shock, and fire, but omits key objects in the ground truth description.}
        \end{tcolorbox}
    \end{minipage}
    \hfill
    \begin{minipage}[t]{0.48\linewidth}
        \includegraphics[width=\linewidth,height=4.9cm]{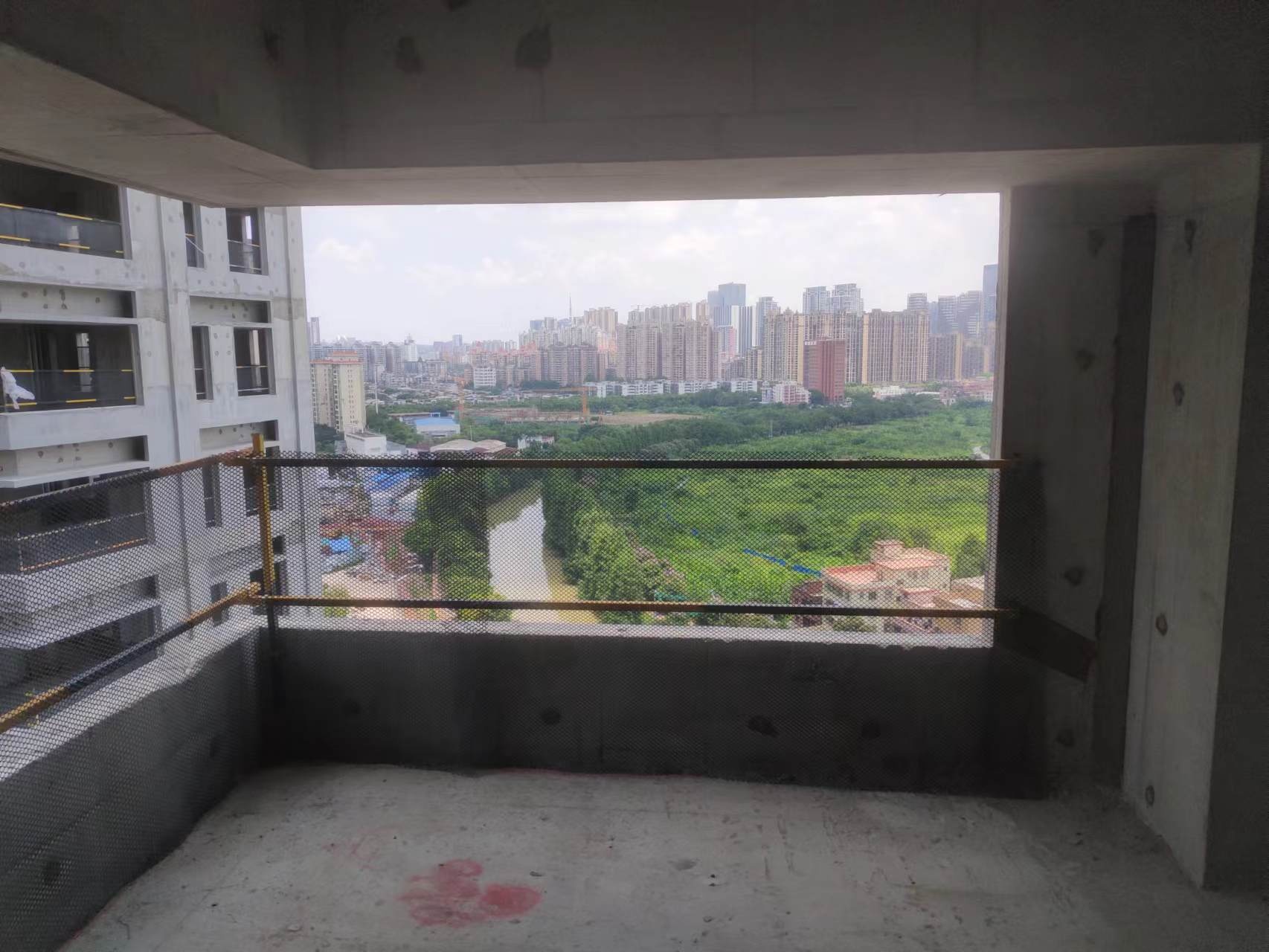}
        \vspace{-2mm}
        \begin{tcolorbox}[title=Task: Open-Ended Description \& Evaluation, colbacktitle=purple!70!black]
            \footnotesize
            \textbf{Ground Truth:} \textit{The \textbf{safety netting} is improperly installed, with \textbf{visible gaps} that could lead to falls from height.}
            \vspace{0.5mm}
            \hrule
            \vspace{0.5mm}
            \textbf{Model Prediction:}
            \textcolor{red!60!black}{There is no visible construction safety hazard in this scene. The area appears secure with a protective barrier in place.} \textcolor{red}{\sffamily [FAIL]}
            \par\vspace{1mm}
            \textbf{\textsc{LLM-as-a-judge Output:}}\par
            \begin{description}[leftmargin=0.1em,itemsep=0pt,topsep=0pt]
                \item[Hazard Detection:] \textbf{0} \textcolor{gray}{(Missed)}
                \item[Quality Score:] \textbf{1/5} \textcolor{gray}{(Poor)}
                \item[Final Score:] \colorbox{red!10}{\textbf{0.0}}
            \end{description}
            \vspace{0.5mm}
            \textit{\textbf{Reasoning:} The predicted description \textbf{fails to detect the hazard} described in the ground truth. It incorrectly states that there is no hazard, making it both \textbf{inaccurate and unhelpful}.}
        \end{tcolorbox}
    \end{minipage}
    \caption{\textbf{Qualitative Examples from \bench.} Two reasoning samples requiring complex analysis of Temporary Electrical Supply and Working At Heights hazards. Key risk factors are bolded.}
    \Description{Two free-form hazard-description examples with construction-site images, ground-truth descriptions, model predictions, and rubric-based judge outputs.}
    \label{fig:fig_reason_sample}
\end{figure*}

Figure~\ref{fig:fig_mcq_sample} and Figure~\ref{fig:fig_reason_sample} show representative examples in the original case-study format. The MCQ samples highlight fine-grained category discrimination with semantically similar distractors, while the open-ended samples show the LLM-as-a-judge scoring fields used for hazard-description evaluation.

\end{document}